\documentclass{article}

\PassOptionsToPackage{numbers, compress}{natbib}

\usepackage[preprint]{neurips_2020}

\usepackage[utf8]{inputenc} 
\usepackage[T1]{fontenc}    
\usepackage{hyperref}       
\usepackage{url}            
\usepackage{booktabs}       
\usepackage{amsfonts}       
\usepackage{nicefrac}       
\usepackage{microtype}      
\usepackage{algorithm}
\usepackage{algorithmic}
\usepackage{amsmath}
\usepackage{wrapfig}
\usepackage{subfigure}
\usepackage{graphicx}
\usepackage{float}
\usepackage{bbm}
\usepackage{multirow}
\usepackage{diagbox}
\usepackage{wrapfig}

\title{Accuracy Prediction with Non-neural Model for Neural Architecture Search}

\author{%
  Renqian Luo\textsuperscript{\rm 1}\thanks{The work was done when the first author was an intern at Microsoft Research Asia.}, Xu Tan\textsuperscript{\rm 2}, Rui Wang\textsuperscript{\rm 2}, Tao Qin\textsuperscript{\rm 2}, Enhong Chen\textsuperscript{\rm 1}, Tie-Yan Liu\textsuperscript{\rm 2}\\
  \textsuperscript{\rm 1}University of Science and Technology of China, \textsuperscript{\rm 2}Microsoft Research Asia\\
  \texttt{lrq@mail.ustc.edu.cn,cheneh@ustc.edu.cn,}\\
  \texttt{\{xuta,ruiwa,taoqin,tyliu\}@microsoft.com}
}

\begin{document}

\maketitle

\begin{abstract}
Neural architecture search (NAS) with an accuracy predictor that predicts the accuracy of candidate architectures has drawn increasing attention due to its simplicity and effectiveness. Previous works usually employ neural network-based predictors which require more delicate design and are easy to overfit. Considering that most architectures are represented as sequences of discrete symbols which are more like tabular data and preferred by non-neural predictors, in this paper, we study an alternative approach which uses non-neural model for accuracy prediction. Specifically, as decision tree based models can better handle tabular data, we leverage gradient boosting decision tree (GBDT) as the predictor for NAS. We demonstrate that the GBDT predictor can achieve comparable (if not better) prediction accuracy than neural network based predictors. Moreover, considering that a compact search space can ease the search process, we propose to prune the search space gradually according to important features derived from GBDT. In this way, NAS can be performed by first pruning the search space and then searching a neural architecture, which is more efficient and effective. Experiments on NASBench-101 and ImageNet demonstrate the effectiveness of using GBDT as predictor for NAS: (1) On NASBench-101, it is 22x, 8x, and 6x more sample efficient than random search, regularized evolution, and Monte Carlo Tree Search~(MCTS) in finding the global optimum; (2) It achieves $24.2\%$ top-1 error rate on ImageNet, and further achieves $23.4\%$ top-1 error rate on ImageNet when enhanced with search space pruning. Code is available at \url{https://github.com/renqianluo/GBDT-NAS}.
\end{abstract}

\section{Introduction}
\label{sec:intro}

\begin{table}
\centering
\caption{Two examples of 4-layer architectures with tabular data representation~(one-hot features) and the corresponding accuracy. `arch' is short for architecture.}
\label{tbl:tablur}
\begin{tabular}{l|c|c}
\toprule
Feature & arch 1 & arch 2\\
\midrule
layer 1 is conv1x1 & 1 & 0 \\
layer 1 is conv3x3 & 0 & 1 \\
layer 2 connects layer 1 & 1 & 1 \\
layer 2 is conv1x1 & 1 & 0 \\
layer 2 is conv3x3 & 0 & 1 \\
layer 3 connects layer 1 & 1 & 0 \\
layer 3 connects layer 2 & 0 & 1 \\
layer 3 is conv1x1 & 0 & 0 \\
layer 3 is conv3x3 & 1 & 1 \\
layer 4 connects layer 1 & 0 & 0 \\
layer 4 connects layer 2 & 0 & 1 \\
layer 4 connects layer 3 & 1 & 1 \\
layer 4 is conv1x1 & 0 & 0 \\
layer 4 is conv3x3 & 1 & 1 \\
\midrule
accuracy (\%) & 92.50 & 93.20 \\
\bottomrule
\end{tabular}
\end{table}

Neural architecture search~(NAS), which aims to automatically find neural network architectures, has been studied and shown its effectiveness in many tasks such as image classification~\cite{nas,nasnet}, object detection~\cite{nasfpn,detnas}, network pruning~\cite{autoslim}, neural machine translation~\cite{evovledtransformer}, text to speech~\cite{seminas}. Representative NAS methods include reinforcement learning based~\cite{nasnet,enas}, evolutionary algorithms based~\cite{amoebanet,darts}, Bayesian methods based~\cite{bayesnas}, gradient based~\cite{nao,darts}, Monte Carlo Tree Search~(MCTS) based~\cite{deeparchitect,mctsnas,lanas} and accuracy predictor based methods~\cite{nao,neuralpredictor,npenas}. Among them, accuracy predictor based approaches, in which an accuracy predictor is used to predict the accuracy of candidate architectures in the search space and save the huge cost induced by training/evaluating these candidate architectures, are simple yet effective. 

Previous works~\cite{perfpred,PNAS,nao,neuralpredictor} usually employ neural network based models such as recurrent neural network~(RNN), convolutional neural network~(CNN), graph convolutional network~(GCN) to build the predictor. Though neural predictors~(i.e., neural network based predictors) have shown promising performance. They are easy to overfit and require delicate design~\cite{nao,neuralpredictor,npenas}. Observing that the discrete representations of architectures are more like tabular data that are preferred by non-neural models~(e.g., tree based models)\footnote{Generally speaking, raw image/text/speech data with spatial or temporal smoothness is preferred by neural networks, graph data is preferred by graph neural networks, while tabular data is preferred by tree based models.},
in this paper, we consider an alternative approach and explore how to build the accuracy predictor based on gradient boosting decision trees (GBDT). Our proposed NAS algorithm with a GBDT based predictor works as follows: 1) We reformulate the general representation of an architecture into one-hot features to make it suitable for GBDT. Given an architecture, we denote the presence or absence of an operator as 1 or 0. For example, we show two architectures and their features in Table~\ref{tbl:tablur}. 2) A GBDT predictor is trained with a few architecture-accuracy pairs. 3) The trained GBDT is used to predict the accuracy of more architectures in the search space and architectures with top predicted accuracy are selected for further evaluation. We call this algorithm GBDT-NAS.

Predicting the accuracy for all the candidate architectures is costly for large search space in real applications~(e.g., search space of MobileNet-V3 is roughly $1e27$), which can take billions of years. Most predictor based methods~\cite{neuralpredictor} choose to randomly sample a number of architectures from the search space to predict rather than all the architectures, resulting in sub-optimal result. Therefore, a compact search space can simplify the search process and help NAS to find better architectures with better sample efficiency. As GBDT is easier to tell the importance/contribution of a feature (i.e., the presence or absence of a candidate network operation), in order to improve sample efficiency, we propose to find not-well-performed candidate operations and prune them from the search space before searching for good architectures. Consequently, we propose to perform NAS by first pruning the search space using the GBDT as a pruner and then searching in the pruned search space using the same GBDT as a predictor, leading to a more efficient and effective NAS method which we call GBDT-NAS with search space search~(GBDT-NAS-S3).

Extensive experiments on NASBench-101 and ImageNet demonstrate the effectiveness of our methods. Specifically, on NASBench-101, GBDT-NAS is 22x, 8x and 6x more sample efficient than random search, regularized evolution~\cite{amoebanet} and MCTS~\cite{mctsnas} respectively. It achieves $24.2\%$ top-1 error rate on ImageNet. Moreover, GBDT-NAS-S3 achieves $23.4\%$ top-1 error rate on ImageNet.

To sum up, our main contributions are listed as follows:
\begin{itemize}
    \item We explore non-neural models (GBDT) as the accuracy predictor to better learn the representation of architectures and perform architecture search~(GBDT-NAS), and show that it leads to better prediction accuracy against neural network based predictors.
    \item We further propose to first prune the search space using GBDT as a pruner and then conduct architecture search using GBDT as a predictor~(GBDT-NAS-S3), which makes the overall search process more efficient and effective.
    \item Experiments on NASBench-101 and ImageNet verify the effectiveness and efficiency of our proposed GBDT-NAS and GBDT-NAS-S3. 
\end{itemize}

\section{Related Works}
\label{sec:rework}
\paragraph{Neural Architecture Search}
\cite{nas} introduces to use reinforcement learning to automatically search neural architecture and brings it to a thriving research area. Lots of works are emerged to explore different search algorithms including reinforcement learning~\cite{nasnet,enas}, evolutionary algorithm~\cite{genetic_cnn,evolvingNN,EA_2017,amoebanet}, Bayesian optimization~\cite{bayesnas}, accuracy prediction~\cite{perfpred,PNAS,nao,neuralpredictor}, gradient based optimization~\cite{nao,darts} and MCTS~\cite{deeparchitect,mctsnas,lanas}. Though all these algorithms have demonstrated effectiveness, accuracy predictor based methods have particularly drawn lots of interests due to its simplicity and effectiveness compared with other algorithms that need careful design and tuning. Accordingly, our proposed method is based on accuracy predictor.

\paragraph{Accuracy Predictor in NAS}
Considering that evaluating candidate architectures raises extremely high cost,~\cite{perfpred} proposes to encode a given architecture to continuous representation via RNN and predict its accuracy rather than really training it to speed up the search process. Further, NAO~\cite{nao} builds the accuracy predictor based on LSTM and fully connected layer to perform gradient based search. However, neural predictors are easy to overfit the architecture-accuracy paired data when applying to different tasks/datasets~\cite{nao,neuralpredictor}, which requires delicate designing and tuning, and implicitly brings additional human efforts.

Recently, GCN is proposed to model architectures and predict the accuracy~\cite{neuralpredictor,npenas}. It points out that NASNet~\cite{nasnet} search space based on Inception~\cite{inceptionv1} backbone for computer vision tasks~\cite{nasnet,enas,darts} is a directed acyclic graph that can be better modeled by GCN. However, in many scenarios, chain-like structure is more commonly used where each layer is only connected to the preceding layer~(e.g., ResNets/Mobilenets in computer vision tasks~\cite{designspace,proxylessnas,efficientnet,onceforall,bignas}, encoder-decoder framework in language tasks~\cite{transformer,dynabert} and speech tasks~\cite{transformertts,seminas}), and the architectures are commonly represented as sequences of discrete symbols from bottom layer to top layer, which are similar to tabular data. Neural predictors are easy to overfit on such structure data and GCN may fail when applying to such chain structure where the adjacent matrix is too sparse and only the diagonal has values. Such tabular data is preferred by non-neural models~(e.g., tree based models) and thus we propose to utilize GBDT as the predictor, which is much simpler and more general to model both graph-like and chain-like structures and can be easily applied to different tasks without much tuning.

\paragraph{Search Space Search}
Search space plays an essential role in the search phase~\cite{rdarts,evanas}. How to get an appropriate search space is critical in NAS. Sampling all the architectures in the search space is expensive and in-efficient, and sampling a small portion leads to sub-optimal. RegNet~\cite{designspace} proposes to progressively prune the large search space via statistical tool~(i.e., empirical distribution function) on a set of randomly sampled architectures to identify the best choice for different factors. Despite the impressive results, the search process heavily relies on human efforts. Specifically, the order of pruning is manually decided by human insights and statistical analyses which is similar to greedy search, and only one factor is considered at each time while different factors may have interactions. Therefore, a natural choice is to leverage models to automatically provide the pruning choices in a global way. GBDT automatically identifies the importance of different features according to some criterion, and is more simple and interpretable than neural networks. We propose to use GBDT to figure out the disappointing candidate operations automatically to prune the search space. Moreover, we can conduct higher-order analysis via combinations of different features during pruning.

LaNAS~\cite{lanas} also proposes to automatically prune the search space recursively. It splits the search space into different sub-regions according to the predicted performance of different regions and different partition thresholds, which relies on the quality of predictor and partition threshold that are not easy to choose. Our method identifies the contribution of each candidate operation to the output which is more simple and efficient.

\section{Non-neural Predictor using GBDT}
\label{sec:gbdtnas}
In this section, we introduce how to use GBDT as accuracy predictor for architecture search, which we call \emph{GBDT-NAS}. In the following paragraphs, we first describe the design of input feature and output value to train a GBDT model, and then formulate our algorithm.  

We describe a discrete neural network architecture as a sequence of tokens from bottom layer to top layer~(e.g., `conv 1x1, conv 3x3, conv 1x1, conv 3x3' to describe a 4-layer neural network, where each position represents the categorical choice for a layer). Considering categorical features may not be a good choice since the relative value of the category ID is meaningless, we convert the category feature into one-hot feature with $O$-dimension, where $O$ is the number of candidate operations and the value of the one-hot feature is `$1$' or `$0$'~(representing whether to use this operation or not). For example, if the candidate operations only contain `conv 1x1 and conv 3x3', then the input feature of the 4-layer architecture demonstrated above is `[1,0,0,1,1,0,0,1]'. Examples of architectures are demonstrated in Table~\ref{tbl:tablur}. The output of GBDT is the accuracy of an architecture, where the target accuracy is normalized to ease model training. We use two ways for normalization: 1) min-max normalization~\cite{nao}, which rescales the values into $[0,1]$, i.e.,  $\frac{y-y_\text{min}}{y_\text{max}-y_\text{min}}$. 2) standardization~\cite{neuralpredictor}, which rescales the accuracy to be zero mean and standard variance, i.e., $\frac{y-y_\text{mean}}{y_\text{std}}$. The training of GBDT aims to minimize the mean squared error between predicted accuracy and target accuracy. We name our GBDT predictor based search algorithm as GBDT-NAS. It contains $T$ iterations and each iteration mainly follows three steps:
\begin{itemize}
    \item \textbf{Train Predictor.} Train the GBDT predictor with $N$ architecture-accuracy pairs.
    \item \textbf{Predict.} Predict the accuracy of $M$ randomly sampled architectures.
    \item \textbf{Validation.} Evaluate $K$ architectures with the top $K$ predicted accuracies. Combine them with the $N$ architecture-accuracy pairs for next iteration.
\end{itemize}
Finally, the architecture with the highest valid accuracy is selected as the final result.

\section{Using GBDT for Search Space Search}
In this section, we leverage the trained GBDT to search the search space by pruning unpromising candidate operations. In this way, searching in the pruned search space is more efficient.

\subsection{Motivation}
Search space is critical for NAS~\cite{random,rdarts}. First, different search spaces have different upper bounds of accuracy that may outweigh the effect of search algorithm. Second, the size of search space affects the effectiveness and efficiency of a search algorithm. 

Ideally, when an accuracy predictor is well trained, one may consider using it to predict the accuracy of all the architectures in the search space~(i.e., set $M$ to be the size of search space). However, when applying to tasks with large search space, traversing all the architectures is time consuming although predicting the accuracy of a single architecture is negligible for GBDT. For example, a commonly used search space~\cite{proxylessnas,neuralpredictor} based on MobileNet-V2~\cite{mobilenetv2} backbone for computer vision tasks consists of more than $1e17$ architectures~\cite{proxylessnas,mobilenetv3,onceforall} which would take millions of years for GBDT to predict on a single CPU and is inefficient. A commonly adopted approach~\cite{neuralpredictor,mctsnas} is to randomly sample a small set~(i.e., $M$) of architectures from the huge search space for prediction, which may lead to sub-optimal result.

\subsection{Pruning Search Space Using GBDT}
We expect to search within a sub-space derived from the large one that contains potentially better architectures, which is more sampling efficient. Considering that GBDT can automatically determine the importance of a feature (the presence or absence of an operation) and can explain the accuracy prediction due to the advantage of tree-based model, in this paper, we leverage the explainable GBDT as the pruner to shrink a search space without human knowledge. A simple way is to use the automatically derived feature importance from the trained GBDT\footnote{The feature importance in GBDT is determined by the average information gain when choosing this feature.}. However, feature importance in GBDT only considers the contribution when training a GBDT model, which may not be entirely consistent with the feature importance in the accuracy of an architecture.

In this paper, we leverage SHapley Additive exPlanation (SHAP) value~\cite{unifiedshap}, which can measure the positive or negative contribution of a feature in GBDT prediction~(i.e., architecture accuracy in the GBDT accuracy predictor) for each sample\footnote{SHAP value~\cite{unifiedshap} is a unified measure of different Shapley values~\cite{shapleyregression,shapleysampling,quantinputinfluence} which reflects the importance of features on the result considering their cooperation and interaction by solving a combined cooperative game theory problem~\cite{shapleyvalue}. It attributes to each feature a value~(real number) showing how it affects the output~(positively or negatively).}. Accordingly, in each iteration, we can get the average SHAP values for each operation in current search space. Then, we select the one with the lowest and extremely negative SHAP value, which implies the most negative contribution on the predicted accuracy, and then prune the search space according the feature. For example, if the average SHAP value of a feature value, layer\_1\_is\_conv1x1=1 is extremely negative~(e.g., $-0.2$) among all the features, then we prune the search space with layer\_1\_is\_conv1x1=1, and then all the architectures in the remaining space have layer\_1\_is\_conv1x1=0~(do not choose conv1x1 at layer\_1). We do this progressively until a certain number of features or all the extremely negative features have been pruned.

Further, since operations in a network may have interactions to cooperatively affect the network accuracy, pruning the search space considering the combinations of several operations is more reasonable and effective. To achieves this, we calculate the interaction SHAP values between any two features, which imply their cooperative contribution to the final accuracy prediction. Then we sort the combinations according to their interaction SHAP values and start from the most negative ones to prune. This can quickly find the most important feature combinations that affect the model output. We name the pruning method that uses SHAP value as first-order pruning shown in Alg.~\ref{alg:1pruning} and that uses interaction SHAP value as second-order pruning demonstrated in Alg.~\ref{alg:2pruning}.

\begin{algorithm}[ht]
\small
\caption{First-Order Pruning}
\label{alg:1pruning}
\begin{algorithmic}[1]
\STATE \textbf{Input}: Trained GBDT accuracy predictor $f$. Current architecture pool $X$. One-hot feature set $F$. Number of features to be pruned $N_{pf}$.
\STATE $Z=\emptyset$.
\STATE $S = SHAP\_Values(f, X)$.
\STATE \text{Sort} $F$ \text{according to} $S$.
\FOR {$l= 1,\cdots, N_{pf}$}
\STATE $fea=F.pop()$.
\STATE $I = \{i|x_i[fea] = 1, x_i \in X\}$.
\STATE $S_{fea} = \sum_{i\in I}S[i,fea] / |I|$.
\IF{$S_{fea}<0$}
\STATE $Z.add(fea)$.
\ENDIF
\ENDFOR
\STATE \textbf{Output}: The pruned feature set $Z$.
\end{algorithmic}
\end{algorithm}

\begin{algorithm}[!h]
\small
\caption{Second-Order Pruning}
\label{alg:2pruning}
\begin{algorithmic}[1]
\STATE \textbf{Input}: Trained GBDT accuracy predictor $f$. Current architecture pool $X$. One-hot feature set $F$. Number of features to be pruned $N_{pf}$.
\STATE $Z=\emptyset$.
\STATE $S = SHAP\_Interaction\_Values(f, X)$.
\STATE $F_2=\{(fea_i,fea_j)|fea_i \in F, fea_j \in F, 0<=i<j<|F|\}$.
\STATE \text{Sort} $F_2$ \text{according to} $S$.
\FOR {$l= 1,\cdots, N_{pf}$}
\STATE $(fea_1, fea_2)=F_2.pop()$.
\STATE $I_{11} = \{i|x_i[fea_1]=1,x_i[fea_2] = 1, x_i \in X\}$.
\STATE $I_{10} = \{i|x_i[fea_1]=1,x_1[fea_2] = 0, x_i \in X\}$.
\STATE $I_{01} = \{i|x_i[fea_1]=0,x_1[fea_2] = 1, x_i \in X\}$.
\STATE $S_{11} = \sum_{i\in I_{11}}S[i,fea_1,fea_2] / |I_{11}|$.
\STATE $S_{10} = \sum_{i\in I_{10}}S[i,fea_1,fea_2] / |I_{10}|$.
\STATE $S_{01} = \sum_{i\in I_{01}}S[i,fea_1,fea_2] / |I_{01}|$.
\IF{$S_{11}<0$}
\STATE $Z.add(fea_1,fea_2)$.
\ELSIF{$S_{10}<0$}
\STATE $Z.add(fea_1)$.
\ELSIF{$S_{01}<0$}
\STATE $Z.add(fea_2)$.
\ENDIF
\ENDFOR
\STATE \textbf{Output}: The pruned feature set $Z$.
\end{algorithmic}
\end{algorithm}

In Alg.~\ref{alg:1pruning}, we first calculate the SHAP values using the trained GBDT predictor $f$ on current architecture-accuracy pairs $X$ in line 3. Then we sort all the features according to the SHAP values in line 4. From line 6 to 10, we get the feature $fea$ with the current most negative SHAP values which indicates its negative contribution to the accuracy and add it the set $Z$ which contains the operations required to be pruned. Alg.~\ref{alg:2pruning} is similar to Alg.~\ref{alg:1pruning} and is different in that we calculate the SHAP interaction values between each two features in line 3, and calculate the SHAP values of all the combinations of two features from line 7 to line 20.

\subsection{GBDT-NAS-S3}
Finally, we formalize our GBDT-NAS enhanced with search space search (GBDT-NAS-S3 for short), which leverages GBDT to first search a good search space (GBDT as a pruner) and then search a good architecture (GBDT as an accuracy predictor, i.e., GBDT-NAS). As shown in Alg.~\ref{alg:GBDT-NAS-S3}, compared to GBDT-NAS, GBDT-NAS-S3 additionally uses the trained GBDT predictor $f$ to perform the search space search by pruning unpromising operations in line 7. If we remove this line, the algorithm degenerates to GBDT-NAS in Sec.~\ref{sec:gbdtnas}.

\begin{algorithm}[ht]
\small
\caption{GBDT-NAS-S3}
\label{alg:GBDT-NAS-S3}
\begin{algorithmic}[1]
\STATE \textbf{Input}: Number of initial architectures $N$. Number of architectures $M$ to predict. Number of top architectures $K$ to evaluate. Number of search iterations $T$. Number of features to prune $N_{pf}$.
\STATE Pruned feature set $Z=\emptyset$.
\STATE Randomly sample $N$ architectures to form $X$.
\STATE Train and evaluate architectures in $X$ and get accuracy numbers $Y$.
\FOR {$l= 1,\cdots, T$}
\STATE Train GBDT $f$ using $X$ and $Y$.
\STATE Prune $N_{pf}$ features from the search space to get the pruned features $Z'$, and $Z=Z \bigcup Z'$.
\STATE Randomly sample $M$ architectures with the constraints $Z$ and get $X_s$.
\STATE Predict the accuracy of the architectures in $X_s$ with $f$ to get $Y_s$.
\STATE Select architectures from $X_s$ with top $K$ predicted accuracy in $Y_s$ and form $X'$.
\STATE Train and evaluate each architecture in $X'$ and get $Y'$.
\STATE $X=X\bigcup X', Y=Y\bigcup Y'$.
\ENDFOR
\STATE \textbf{Output}: The architecture within $X$ with the best accuracy.
\end{algorithmic}
\end{algorithm}

\section{Experiments}
We demonstrate the effectiveness of our proposed methods through experiments on two datasets: NASBench-101~\cite{nasbench101} and ImageNet. Since the search space of NASBench-101 is quite small, we only evaluate GBDT-NAS. We evaluate both GBDT-NAS and GBDT-NAS-S3 on ImageNet, which has much larger search space.

\subsection{Analysis of GBDT as Predictor}
We first analyze the performance of GBDT as predictor using NASBench-101. NASBench-101 is a dataset for evaluating NAS algorithms, which eliminates the efforts of training and evaluating candidate architectures. It defines a narrow search space containing only $423k$ CNN architectures and each architecture has been trained and evaluated on CIFAR-10 following exactly the same pipeline and setting. Thus, one can get architecture-accuracy pairs effortless via directly querying from the dataset, and use them to quickly evaluate a NAS algorithm and fairly compare it with other algorithms. The search space is graph-like structure following~\cite{nasnet,enas,darts} which involves candidate connections between different nodes besides operations. We follow the encoding guide by the authors to represent the architecture in a sequence~\cite{nasbench101}. For each node, we use its adjacent vector concatenated with its operation to represent it. Standardization is used to rescale the accuracy when training predictors.

We randomly sample $1100$ architecture-accuracy pairs from the dataset. $1000$ of them are used as training set and the remaining $100$ pairs are used as test set. We train a GBDT model based on LightGBM~\cite{lightgbm} with $100$ trees and $31$ leaves per tree. We also evaluate LSTM, GCN and Transfomer based accuracy predictors as baselines. For LSTM based predictor, we follow NAO~\cite{nao} and use a single layer LSTM of hidden size $16$ followed by two fully connected layers of hidden size $64$. For GCN based accuracy predictor, we follow~\cite{neuralpredictor} with a 3-layer GCN of hidden size $144$ followed by a fully connected layer of hidden size $128$. For Transformer based accuracy predictor, we follow~\cite{transformer} and use a 4-layer Transformer model.
\begin{table}
\centering
\caption{Pairwise accuracy of different predictors. Transformer* and LSTM* represent the setting that we carefully tune the hyper-parameters~(e.g., depth, width, dropout, weight decay, normalization) of the two predictors on NASBench-101, which consumes a lot of human efforts. For GBDT, we directly use the default setting in LightGBM without much tuning.}
\label{tbl:nasbench:acc}
\begin{tabular}{lcc}
\toprule
Model  & Train Acc. (\%) & Test Acc. (\%)\\
\midrule
Transformer & 92 & 58\\
Transformer* & 88 & 65\\
LSTM & 96 & 73 \\
LSTM* & 87 & 80\\
GCN & 85 & 80\\
\midrule
GBDT & 88 & \textbf{82}\\
\bottomrule
\end{tabular}
\end{table}

All the models are trained on the same training set and tested on the same test set described above. To evaluate the predictors, we compute the pairwise accuracy following~\cite{nao} on the held out test set via $\frac{\sum_{x_1\in X, x_2\in X}\mathbbm{1}_{f(x_1)\geq f(x_2)}\mathbbm{1}_{y_1\geq y_2}}{|X|(|X|-1)/2}$, where $\mathbbm{1}$ is the 0-1 indicator function. We run each setting for $100$ times and report the average results in Table~\ref{tbl:nasbench:acc}. 

It shows that the first two neural network based predictors (Transformer and LSTM) are easy to overfit. Even after careful hyper-parameter tuning, Transformer still shows severe overfitting, and LSTM has some improvements but still archives lower test accuracy than GBDT. GCN achieves good performance as we use the already tuned setting for NASBench by~\cite{neuralpredictor}. Non-neural predictor (GBDT) achieves better prediction accuracy~($82\%$) than neural networks~(Transformer, LSTM, GCN). Note that graph-like architectures can not only be well modeled by GCN, but also be well modeled by GBDT which treats them as tabular data. However, training a GBDT model is much faster than training neural models and do not require much design of the neural predictor architecture or tuning of the hyper-parameters.

\subsection{NASBench-101}
\paragraph{Setup} 
We then evaluate the performance of GBDT-NAS on NASBench-101. We mainly compare our method with several baselines: random search, regularized evolution~\cite{amoebanet}, MCTS~\cite{mctsnas} and GCN~\cite{neuralpredictor}. Since the search space of NASBench-101 is small, we set $M$ to be the size of the whole search space and search for only 1 iteration~(i.e., $T=1$). We run each algorithm for $100$ times.

\begin{figure}[htbp]
\vspace{-10pt}
\centering
\includegraphics[width=0.8\columnwidth]{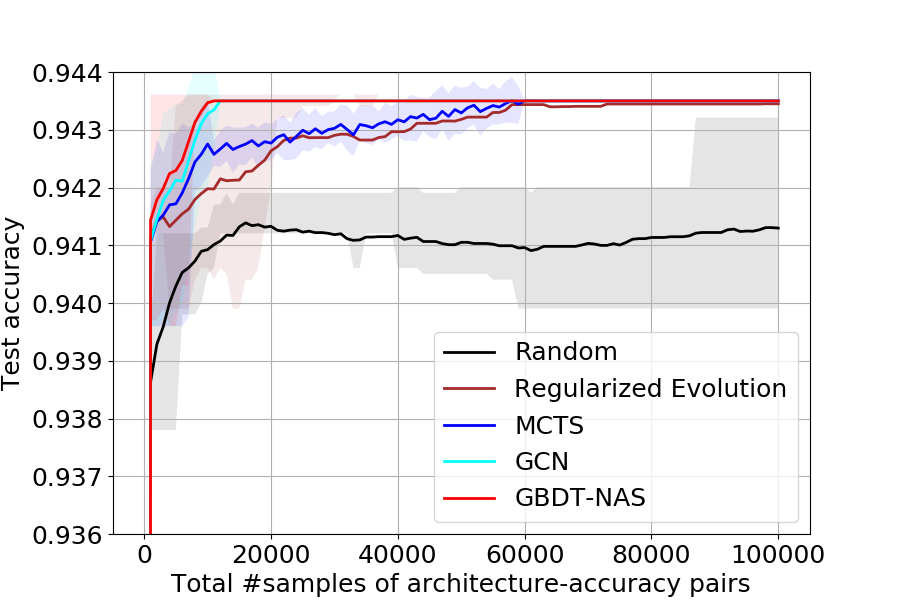}
\caption{Results of different methods on NASBench-101.}
\label{fig:nasbench}
\end{figure}

\paragraph{Results} Following the guidelines suggested in NASBench-101 original publication, we plot the best test accuracy discovered with respect to the number of architecture-accuracy samples used by the algorithm in Fig.~\ref{fig:nasbench}. GBDT-NAS is on average using
\textbf{22x}, \textbf{8x} and \textbf{6x} fewer architecture-accuracy samples than Random Search, Regularized Evolution, and MCTS respectively to find the global optimal. Though the search space in NASBench-101 is graph-like and well modeled by GCN, GBDT shows comparable performance by encoding the architectures as tabular data. Moreover, neural predictors required delicate design~(e.g., depth, width, dropout) to avoid overfitting~\cite{neuralpredictor} while GBDT can easily fit the data without much tuning.

\subsection{ImageNet}
\begin{table}[htbp]
\centering
\small
\begin{tabular}{lccc}
\toprule
Model/Method                      & Top1/Top5 Err. & Params & FLOPS \\ 
\midrule
MobileNetV2~\cite{mobilenetv2}     & 25.3/-        & 6.9M  & 585M \\
ShuffleNetV2~\cite{shufflenet}  & 25.1/-  & $\sim$ 5M & 591M  \\
\midrule
NASNet-A~\cite{nas}               & 26.0/8.4  & 5.3M      & 564M \\
AmoebaNet-A~\cite{amoebanet}      & 25.5/8.0  & 5.1M      & 555M \\
DARTS~\cite{darts}                & 26.9/9.0  & 4.9M      & 595M \\
AlphaX~\cite{mctsnas}            & 24.5/7.8  & 5.4M      & 579M\\
PC-DARTS~\cite{pcdarts}           & 24.2/7.3  & 5.3M      & 597M\\
Efficienet-B0~\cite{efficientnet} & 23.7/6.8 & 5.3M      & 390M \\
DenseNAS~\cite{densenas}          & 23.9/-   & -         & 479M \\
CARS~\cite{cars}                  & 24.8/7.5 & 5.1M      & 591M \\
PC-NAS~\cite{pcnas}               & 23.9/-   & 5.1M      & - \\
\midrule
Random Search                    & 25.2/8.0     & 5.1M   & 578M \\
NAO~\cite{nao}                   & 24.5/7.8     & 6.5M   & 590M \\
ProxylessNAS~\cite{proxylessnas}  & 24.0/7.1     & 5.8M   & 595M\\
Manual Pruning                  & 24.1/7.0     & 6.1M   & 550M \\
LaNAS~\cite{lanas}            & 25.0/7.7       & 5.1M   & 570M \\
\midrule 
GBDT-NAS                      & 24.2/7.1   & 5.8M      & 588M \\
GBDT-NAS-S3 (1st)   & 23.8/6.9   & 5.6M      & 572M \\
GBDT-NAS-S3 (2nd)  & \textbf{23.4}/\textbf{6.7} & 5.7M & 588M \\
\bottomrule
\end{tabular}
\caption{Performances of different methods on ImageNet dataset. For NAO, we use the open source code and search on the same search space used in this paper. We run ProxylessNAS by optimizing accuracy without latency for fair comparison. `1st' and `2nd' indicate using first-order and second order SHAP values for pruning respectively.}
\label{tbl:ImageNet}
\end{table}
\paragraph{Setup}
We adopt the search space used in ProxylessNAS~\cite{proxylessnas} which is based on MobileNetV2 backbone and is chain-like. It is not suitable to be modeled by GCN as the adjacent matrix is too sparse where only diagonal has values. Such chain-like structure is more like tabular data and preferred by GBDT rather than neural predictors. We search the operation of each layer and the candidate operations include mobile inverted bottleneck convolution layers~\cite{mobilenetv2} with kernel size in $\{3, 5, 7\}$ and expansion ratio in $\{3, 6\}$, and identity layer to perform elastic network depth, which yields totally $(3\times2+1)^{21}\approx5e17$ candidate architectures. 

During search, we split out $5000$ images from the training set for validation. Since training on ImageNet is too costly, to reduce the resource required, we adopt the commonly used weight-sharing method to train the candidate architectures in a super-net~\cite{oneshot,onceforall}. We train the super-net containing all the candidates for $20000$ steps at each iteration with a batch size of $512$. 

The GBDT is trained with the same hyper-parameters used in NASBench-101 experiment. Min-max normalization is applied to normalize the accuracy numbers for GBDT. We use $N=1000,M=5000,K=300,T=3$ for evaluating both GBDT-NAS and GBDT-NAS-S3 as described in Alg.~\ref{alg:GBDT-NAS-S3}, according to preliminary study considering both effectiveness and efficiency. Since the search space contains $7$ candidate operations and $21$ layers, the number of features for an architecture is $7\times 21=147$. Specifically in GBDT-NAS-S3, we prune $N_{pf}=20$ features at each iteration to quickly narrow the space. The search runs for $1$ day on 4 NVIDIA V100 GPUs.

We limit the FLOPS of the discovered architecture to be less than $600$M for fair comparison with other works~\cite{nasnet,amoebanet,darts,efficientnet,pcdarts,pdarts} and train it for $300$ epochs. We implement random search as a baseline by randomly sampling $2000$ architectures and training them using the super-net. The one with the best validation accuracy is selected for final evaluation. We also implement manual pruning to perform search space search as a baseline where we sequentially prune disappointing operations by doing statistics on architectures with certain operations similar to~\cite{designspace}.
\paragraph{Results} We list the results in Table~\ref{tbl:ImageNet}. We mainly compare our results to works sharing the same search space, and will apply it to very recently proposed better search space in~\cite{onceforall,bignas,fbnetv2,fbnetv3} that have better accuracy. Results of our experiments are averaged across $5$ runs. We can see that our proposed methods all demonstrate promising results. When using GBDT only as accuracy predictor, GBDT-NAS achieves $24.2\%$ error rate. This shows that GBDT fits the chain-like architecture structure well. Further, when enhanced with search space search, GBDT-NAS-S3 achieves more improvements. Second-order pruning with $23.4\%$ error rate outperforms first-order pruning with $23.8\%$ error rate, demonstrating the effectiveness of considering combinations of feature interactions during search. We will try to use higher-order SHAP value to prune the search space in the future. Compared to other NAS works, our GBDT-NAS-S3 achieves better top-1 error rate under the $600$M FLOPS constraints.

\section{Analysis}
In this section, we study the hyper-parameters in GBDT-NAS. We mainly study $N,K$ which respectively stands for the number of architecture-accuracy pairs to train the GBDT predictor, and the number of architectures to select for further validation.

\begin{figure}[htbp]
\centering
\subfigure[]{
		\label{fig:study_n}
		\includegraphics[width=0.48\columnwidth]{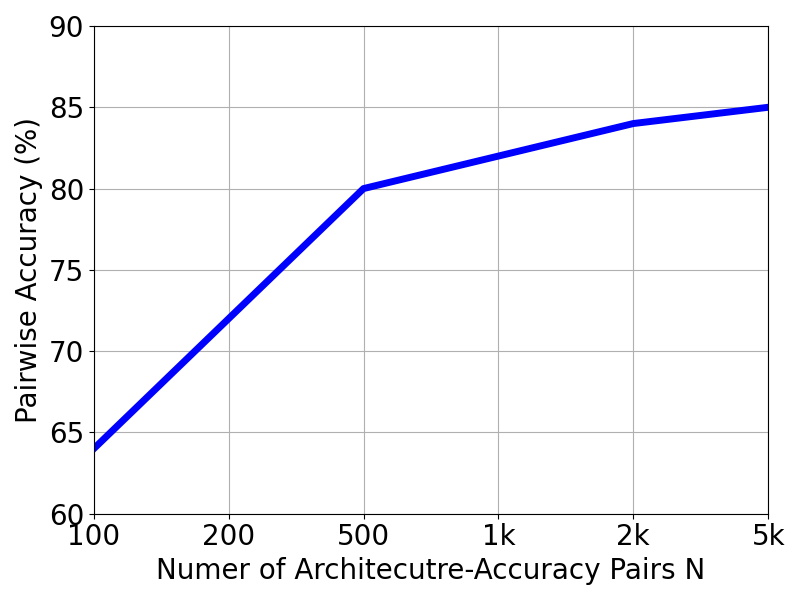}
	}
	\subfigure[]{
		\label{fig:study_k}
		\includegraphics[width=0.48\columnwidth]{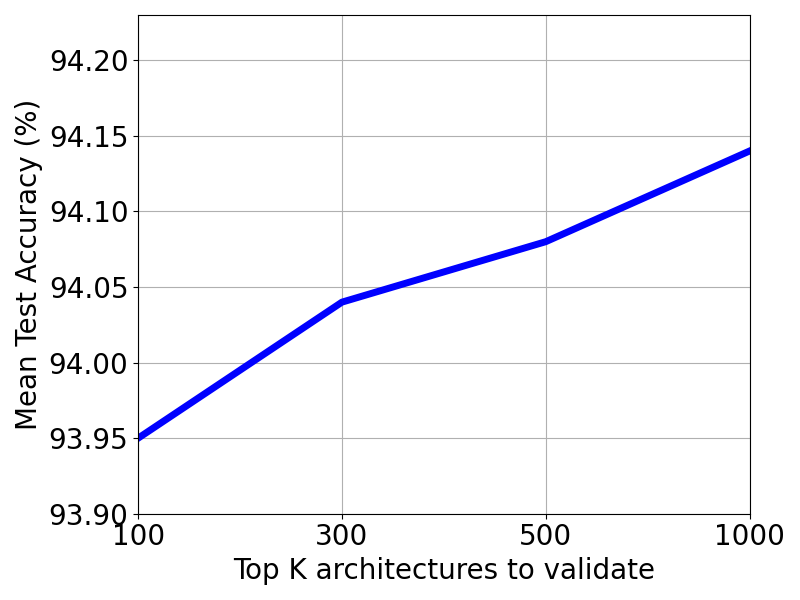}
	}
\caption{(a) Pairwise accuracy of GBDT predictor under different $N$. (b) Mean test accuracy of discovered architecture on NASBench-101 under different $K$.}
\end{figure}

\subsection{Study of Hyper-parameter $N$}
Since GBDT predictor trains on $N$ architecture-accuracy pairs, the number $N$ is critical to the effect of the predictor. Small $N$ may result in bad accuracy and large $N$ leads to more resources required. Following the experiments of evaluating the accuracy predictor, we train the GBDT on $N$ architecture-accuracy pairs queried from NASBench-101 dataset, and measure the pairwise accuracy of the GBDT predictor on a held-out set containing $100$ architecture-accuracy pairs. We vary the value of $K$ and evaluate on NASBench-101. Results are depicted in Fig.~\ref{fig:study_k}. We can see that, when only a small number of architectures with top predicted accuracy are validated, the final discovered architecture shows a moderate test accuracy. With more architectures being evaluated, the discovered architecture achieves better test accuracy.

\subsection{Study of Hyper-parameter $K$}
Since the predictor is not $100\%$ accurate, we cannot completely rely on the prediction to rank all the architectures. Therefore we need to further evaluate the top $K$ architectures by really training and validating them on the valid dataset~(querying the valid accuracy of these architectures in NASBench-101). Small $K$ may potentially miss some well performing architectures that predicted to be bad by the predictor incorrectly, and large $K$ leads to more resource required. We set $N=1000$ and vary the value of $K$ in GBDT-NAS and evaluate on NASBench-101. Results are depicted in Fig.~\ref{fig:study_k}. We can see that, when only a small number of architectures with top predicted accuracy are validated, the final discovered architecture shows a moderate test accuracy. With more architectures are validated, the discovered architecture achieves better test accuracy. This implies that since the predictor is not 100\% accurate, we cannot fully rely on its prediction to return the one with the best predicted accuracy. We need to select a number of architectures with the top predicted accuracies for further evaluation~(training and validation) and return the one with the highest validated accuracy as the discovered one.

\subsection{Discovered Architectures and Analysis of Pruning the Search Space via GBDT}
We show the architectures discovered for ImageNet, and conduct analysis on the effect of using GBDT for search space pruning.

\begin{figure}[htbp]
\centering
\includegraphics[width=.8\columnwidth]{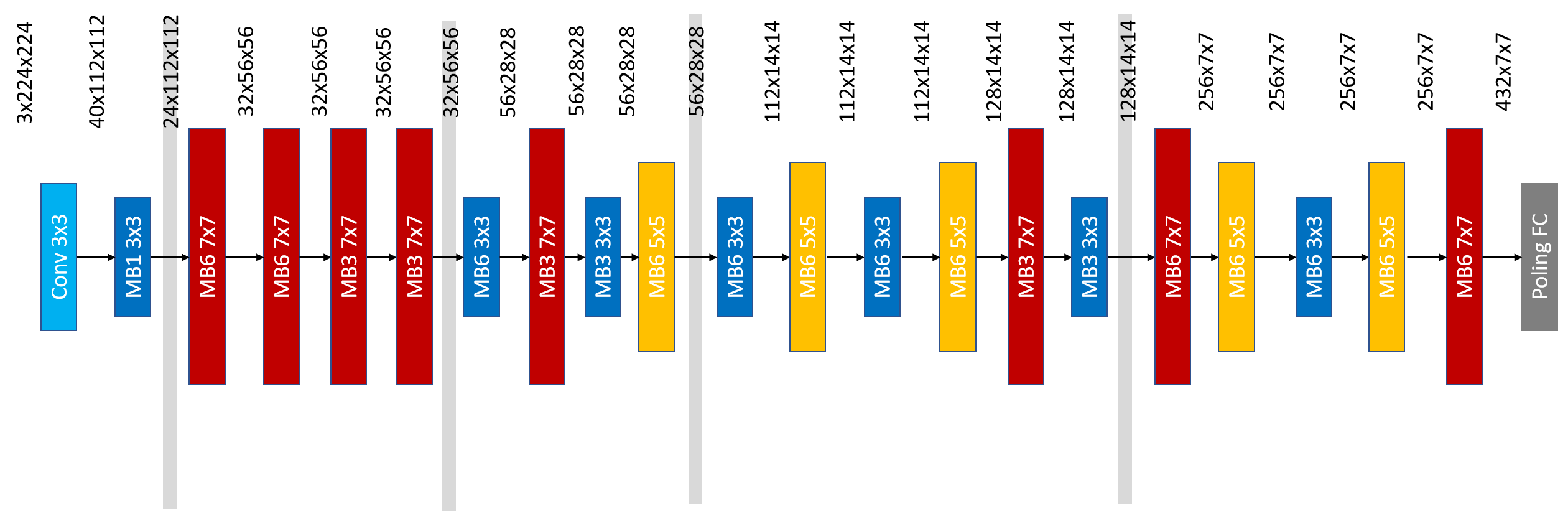}
\caption{Architecture discovered by GBDT-NAS.}
\label{fig:arch1}
\end{figure}

\begin{figure}[htbp]
\centering
\includegraphics[width=.8\columnwidth]{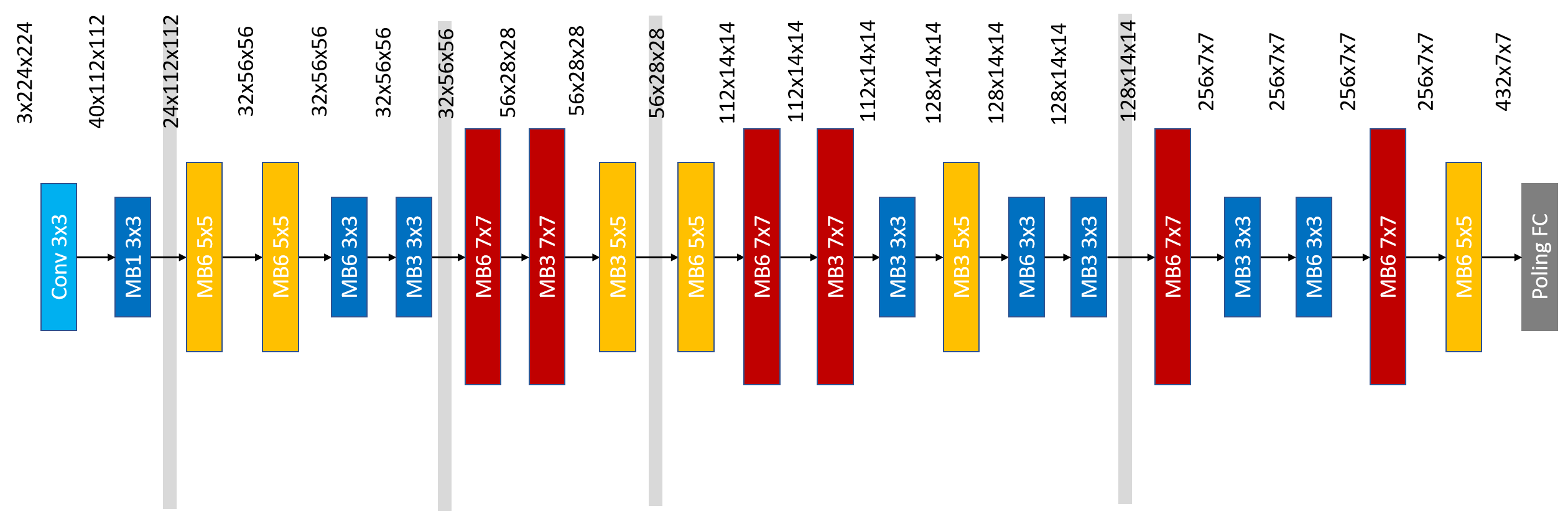}
\caption{Architecture discovered by GBDT-NAS-S3 (first-order pruning).}
\label{fig:arch2}
\end{figure}

\begin{figure}[htbp]
\centering
\includegraphics[width=.8\columnwidth]{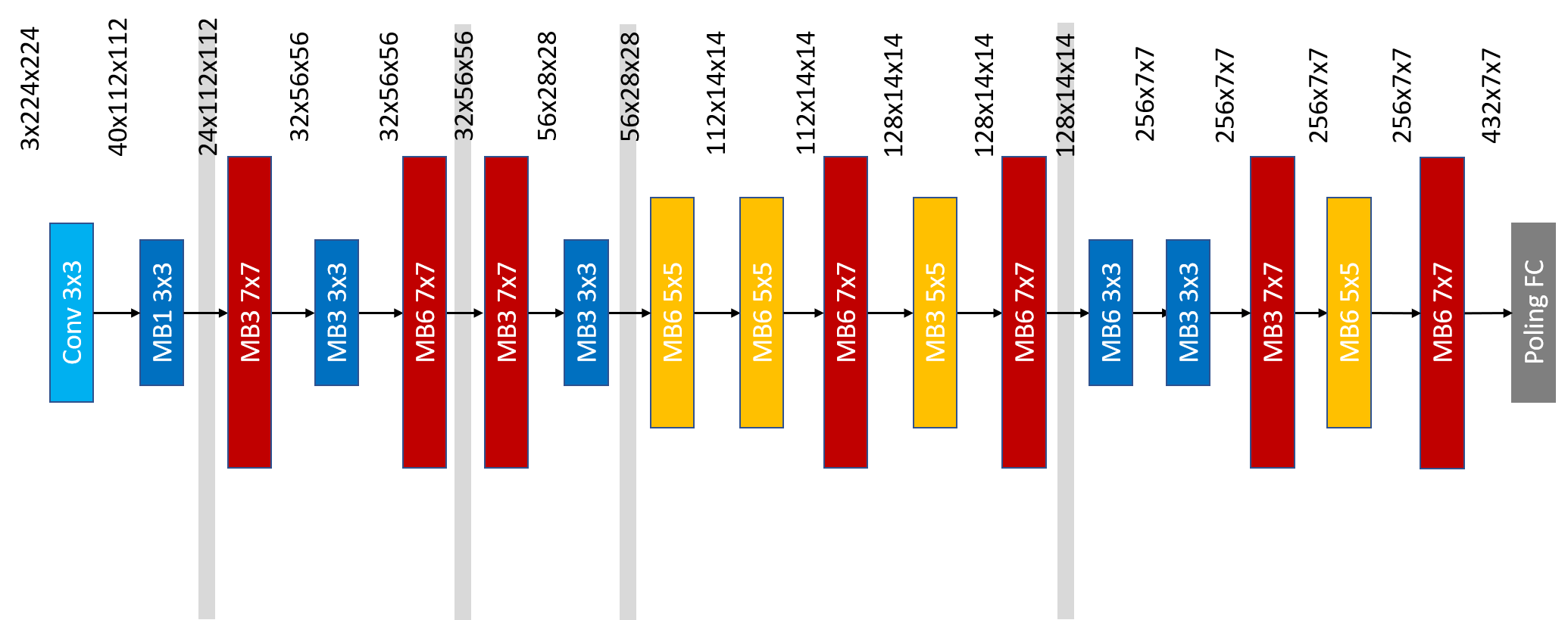}
\caption{Architecture discovered by GBDT-NAS-S3 (second-order pruning).}
\label{fig:arch3}
\end{figure}

We first plot the architectures for ImageNet discovered by GBDT-NAS, GBDT-NAS-S3 (first-order pruning) and GBDT-NAS-S3 (second-order pruning) in Fig.~\ref{fig:arch1}, Fig.~\ref{fig:arch2} and Fig.~\ref{fig:arch3} respectively.

\begin{figure}[htbp]
\centering
\includegraphics[width=0.8\columnwidth]{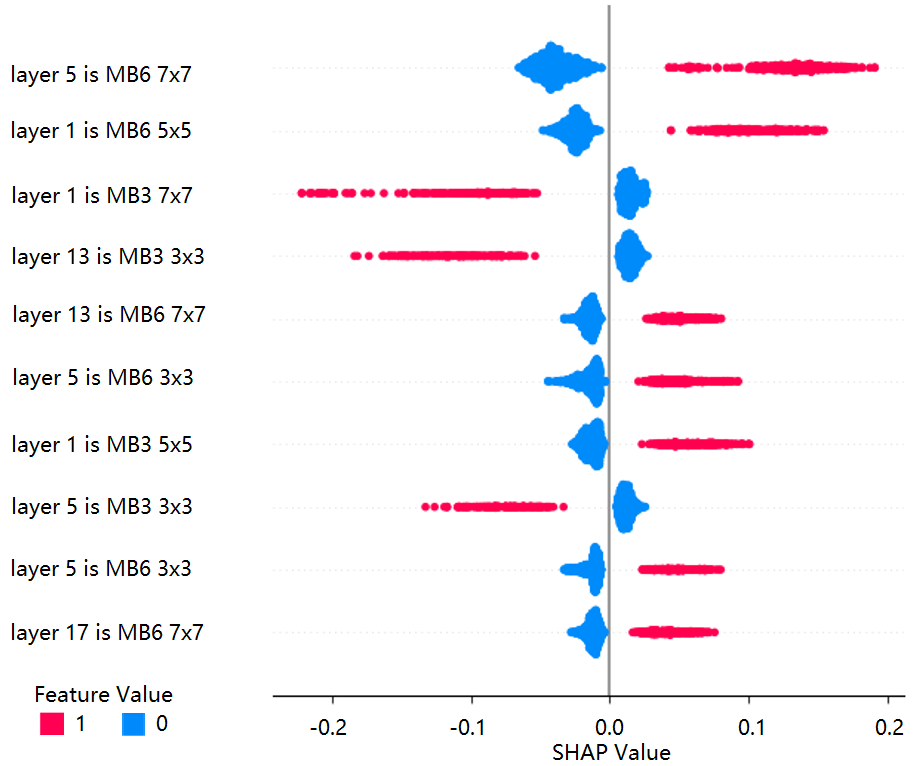}
\caption{SHAP values for several features.}
\label{fig:shap}
\end{figure}

Then, to demonstrate how we perform pruning using SHAP values, we visualize the SHAP values of some features using the official tool\footnote{\url{https://github.com/slundberg/shap}} in Fig.~\ref{fig:shap}. Notice that the colored area contains multiple data points~(architectures). Taking \textbf{`layer 1 is MB3 7x7'} as an example, the SHAP value of this feature is extremely negative when the feature value is `$1$', which indicates that using a `MB3' layer with kernel size $7$ at layer 1 usually has bad accuracy. So we prune this feature and the following sampling process will not sample architectures that use `MB3 7x7' at layer 1. The architecture discovered by GBDT-NAS in Fig.~\ref{fig:arch1} uses `MB3 7x7' at `layer 1'~(`layer 1' starts from the layer right after the first gray bar, while the first two layers `Conv 3x3' and `MB1 3x3' before the bar are fixed as stem layers~\cite{proxylessnas}), which results in the final test error rate of $24.2\%$. However, the two architectures discovered by GBDT-NAS-S3 in Fig.~\ref{fig:arch2} and Fig.~\ref{fig:arch3} do not choose this operation at `layer 1' as the operation is pruned due to its negative effect to the prediction determined by the SHAP value during the search. And these two architectures show better test error rate~($23.8\%$ and $23.5\%$) against the one by GBDT-NAS.

\begin{figure}[htbp]
\centering
\includegraphics[width=0.8\columnwidth]{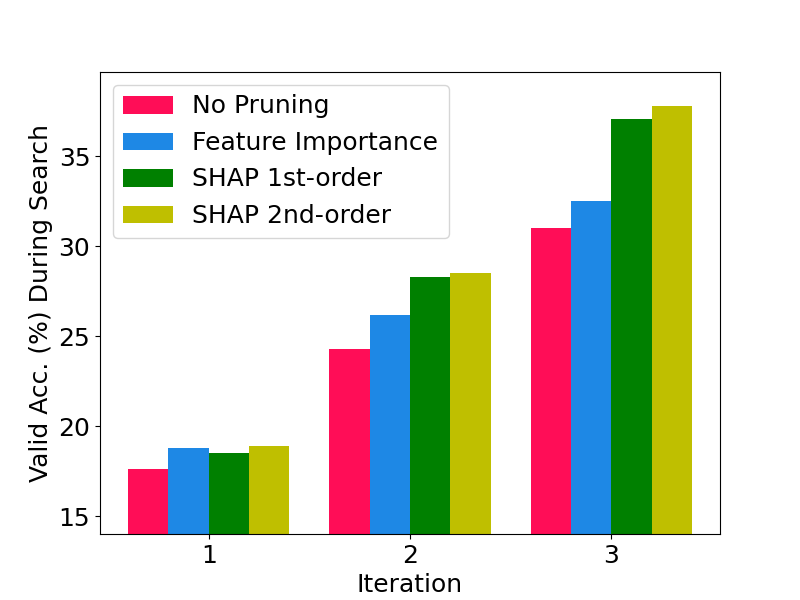}
\caption{Average valid accuracy of the architectures sampled from search space pruned by different methods, evaluated with the shared weights during the search phase.}
\label{fig:acc}
\end{figure}

Further, to demonstrate the effectiveness of the pruning methods during the search phase, we compare the average valid accuracies of architectures sampled from the search space with different pruning methods at each iteration~(We totally run for $T=3$ iterations) in Fig.~\ref{fig:acc}. 

The baseline uses no pruning method~(i.e., GBDT-NAS, as shown in the red bar). We also show the results of pruning according to feature importance determined by GBDT for comparison. Specifically, we sort the features by their feature importance, and then prune the features in sequence. For each feature, we first calculate the average valid accuracies of architectures with and without the feature. Then the feature is pruned if the average valid accuracy of architectures with the feature is lower than the ones without the feature and the gap exceeds a certain margin~(e.g., $1\%$). 

Note that for each single method, the valid accuracy increases with more iterations since the super-net is trained to be better. At each iteration, compared to baseline without pruning, pruned search spaces show higher accuracy. Meanwhile SHAP value based pruning methods outperform the feature importance based method.  This demonstrates that our GBDT based pruning indeed finds better sub-space. Moreover, the gap between SHAP value based pruning methods and baseline is increasing, implying that the sub-space after each iteration is becoming better.

\section{Conclusion and Future Work}
In this paper, considering the tabular data representation of architectures which is preferred by non-neural models, we introduce GBDT into neural architecture search and develop two NAS algorithms: GBDT-NAS and GBDT-NAS-S3. In GBDT-NAS, we use GBDT as accuracy predictor to predict the accuracy of architectures. We further enhance GBDT-NAS with search space search and propose GBDT-NAS-S3, which additionally leverages learned information by GBDT to prune the search space. Experiments on NASBench-101 and ImageNet demonstrate the effectiveness of both methods. In our future work, we plan to use GBDT to search in more general search space and on more complicated tasks.

\section*{Acknowledgements}
We sincerely thank Guolin Ke for his valuable comments and suggestions.

\bibliography{main_arxiv}
\bibliographystyle{plain}

\clearpage
\centerline{\Huge{Appendix}}

\setcounter{section}{0}

\section{The SHAP Value based Pruning Algorithm}
In this section, we describe the pruning algorithm using SHAP value. We get the SHAP value of some architecture-accuracy pairs sampled from the search space. Then we sort the features according to the SHAP values. We start pruning from the most negative one. Since the one-hot feature represents using or not using the corresponding operation, we only prune the features with value `1' (indicating that using this operation may lead to inferior prediction). Feature with value `0' will not be pruned since by default the operation is to be sampled. For example, by default `conv1x1' is in the candidate operations and will be sampled. When `layer\_1\_is\_conv1x1=1' has a very negative SHAP value which means using `conv1x1' at `layer\_1' will lead to inferior prediction, we prune the `layer\_1\_is\_conv1x1' from the search space and `conv1x1' will not be sampled to be the operation of `layer\_1'. However, when `layer\_1\_is\_conv1x1=0' has a very negative SHAP value which means not using `conv1x1' at `layer\_1' will lead to inferior prediction, we do nothing since by default `conv1x1' will be sampled to be the operation of `layer\_1' in other cases. We give the first-order pruning algorithm in Alg.~\ref{alg:1pruning} and second-order pruning algorithm in Alg.~\ref{alg:2pruning}.

\begin{algorithm}[ht]
\small
\caption{First-Order Pruning}
\label{alg:1pruning}
\begin{algorithmic}[1]
\STATE \textbf{Input}: Trained GBDT performance predictor $f$. Current architecture pool $X$. One-hot feature set $F$. Number of features to be pruned $N_{pf}$.
\STATE $Z=\emptyset$.
\STATE $S = SHAP\_Values(f, X)$.
\STATE Sort $F$ according to $S$.
\FOR {$l= 1,\cdots, N_{pf}$}
\STATE $fea=F.pop()$.
\STATE $I = \{i|x_i[fea] = 1, x_i \in X\}$.
\STATE $S_{fea} = \sum_{i\in I}S[i,fea] / |I|$.
\IF{$S_{fea}<0$}
\STATE $Z.add(fea)$.
\ENDIF
\ENDFOR
\STATE \textbf{Output}: The pruned feature set $Z$.
\end{algorithmic}
\end{algorithm}

\begin{algorithm}[ht]
\small
\caption{Second-Order Pruning}
\label{alg:2pruning}
\begin{algorithmic}[1]
\STATE \textbf{Input}: Trained GBDT performance predictor $f$. Current architecture pool $X$. One-hot feature set $F$. Number of features to be pruned $N_{pf}$.
\STATE $Z=\emptyset$.
\STATE $S = SHAP\_Interaction\_Values(f, X)$.
\STATE $F_2=\{(fea_i,fea_j)|0<=i<j<|f|\}$.
\STATE Sort $F_2$ according to $S$.
\FOR {$l= 1,\cdots, N_{pf}$}
\STATE $(fea_1, fea_2)=F_2.pop()$.
\STATE $I_{11} = \{i|x_i[fea_1]=1,x_i[fea_2] = 1, x_i \in X\}$.
\STATE $I_{10} = \{i|x_i[fea_1]=1,x_1[fea_2] = 0, x_i \in X\}$.
\STATE $I_{01} = \{i|x_i[fea_1]=0,x_1[fea_2] = 1, x_i \in X\}$.
\STATE $S_{11} = \sum_{i\in I_{11}}S[i,fea_1,fea_2] / |I_{11}|$.
\STATE $S_{10} = \sum_{i\in I_{10}}S[i,fea_1,fea_2] / |I_{10}|$.
\STATE $S_{01} = \sum_{i\in I_{01}}S[i,fea_1,fea_2] / |I_{01}|$.
\IF{$S_{11}<0$}
\STATE $Z.add(fea_1,fea_2)$.
\ELSIF{$S_{10}<0$}
\STATE $Z.add(fea_1)$.
\ELSIF{$S_{01}<0$}
\STATE $Z.add(fea_2)$.
\ENDIF
\ENDFOR
\STATE \textbf{Output}: The pruned feature set $Z$.
\end{algorithmic}
\end{algorithm}

\end{document}